# Accurate indoor mapping using an autonomous unmanned aerial vehicle (UAV)

Lachlan Dowling, Tomas Poblete, Isaac Hook, Hao Tang, Ying Tan, Will Glenn, Ranjith R Unnithan

*Abstract*—An autonomous indoor aerial vehicle requires reliable simultaneous localization and mapping (SLAM), accurate flight control, and robust path planning for navigation. This paper presents a system level combination of these existing technologies for 2D navigation. An Unmanned aerial vehicle (UAV) called URSA (Unmanned Recon and Safety Aircraft) that can autonomously flight and mapping indoors environments with an accuracy of 2 cm was developed. Performance in indoor environments was assessed in terms of mapping and navigation precision.

*Index Terms*—Unmanned aerial vehicles (UAVs), Autonomous obstacle avoidance, Indoor mapping.

## I. Introduction

Autonomous indoor mapping using mobile robots or any mobile device is a necessary tool where human access to different places can be limited due to space constraints or security reasons [1]. For emergency crew, it is a necessary tool for planning the best and more secure mode to operate at dangerous sites [2, 3]. This, by reducing the incidence of personnel exposed to hazardous or tedious conditions [4]. Also, in places where the space or GPS information is reduced, autonomous navigation is a powerful tool where mapping can give information about the surrounding area. Autonomous mapping mechanisms have been implemented in different fields such as architecture and building managing [2], for unknown building interiors mapping [5] and for abandoned mines mapping [6]. Based on their low cost, high spatial resolution and versatility [7], the interest of using Unmanned aerial vehicles (UAVs) have recently increased among several industries and investigation fields, such as traffic monitoring [8], urban planning [9], civil security applications [10, 11], forestry and agriculture [12, 13]. Conventional UAVs flown at high altitudes, regulate their position by continuously monitoring and merging data from an inertial measurement unit (IMU) and a global positioning system (GPS). Nevertheless, in confined areas such as cities, forest and buildings it is not possible to regulate UAV altitude and position because the GPS information is weak or not reliable. In addition, to achieve a complete autonomous flight-controlled system it is necessary to map the UAV's surrounding with high precision and accuracy in order to identify obstacle-free trajectories [14].

R. Unnithan, Ting Tao and Tomas Poblete are with the University of Melbourne department of Electrical and Electronics Engineering (e-mail: r.ranjith@unimelb.edu.au; yingt@unimelb.edu.au; tomas.poblete@unimelb.edu.au)

L. Dowling, I. Hook and H. Tang were Masters students at the University of Melbourne (e-mail: dowlingl@student.unimelb.edu.au; ihook@student.unimelb.edu.au; hao.tang1994@gmail.com).

Will Glenn, belongs to the Metropolitan Fire Brigade.

In regard, the components for an autonomous controlled UAV, consist in a combination of localization and mapping (SLAM), sensor fusion and navigation algorithms. This paper presents the implementation and evaluation of an Unmanned Recon and Safety Aircraft (U.R.S.A.). This prototype can enter to an indoor environment, navigate autonomously, and generate a map of the area and live video feed by using a 2D laser scan data for navigation. Results are presented in terms of both qualitative assessments of navigation performance in different environments, as well as a quantitative comparison of floor map measurements against ground truth data. The paper is organized as follows: Section II, presents a review of the existing technologies. In section III and IV, the design of the URSA prototype and its navigation systems derived from the previous section, are described. The experiments carried out and the results obtained are reported in section V. Finally, the main conclusions and future related works are mentioned in section VI.

## II. Existing technologies

In this section, a literature review about the existing autonomous UAV Systems, SLAM algorithms, flight controlling and navigation is discussed. This section is followed by presenting the architecture of the UAV with implemented methodologies in sections III and IV.

### A. Autonomous UAV systems

A pivotal systems-level contribution towards a complete indoor autonomous UAV was reported by Fei, et al. [15]. The demonstrated UAV combined two-stage approach where the UAV explored its environment using a laser scanner to avoid obstacles and then an optical flow for basic position estimation. The collected laser scan data obtained during the first stage was converted into a map offline using the FastSLAM algorithm. Then, the built map was used as a guide for future flights. In addition, Zeng, et al. [16] showed to localize and map by replacing a laser scanner with a laser source which is detected by a monocular camera. The proposed methodology was based on tracking points of interest overtime, by matching scans at a significantly faster rate than the FastSLAM algorithm with an odometer accuracy of 2% after correction. Nevertheless, as the methodology identifies points of interest only, its utility in constructing a full map is limited. Kumar, et al. [17] demonstrated two orthogonal laser scanners and simple point-to-point scan matching method for navigation and created a fast and efficient indoor mapping and localization solution with translation of around 3 cm. This methodology allowed computation over embedded computers, removing the need for a base station.

## B. Simultaneous localization and mapping (SLAM)

SLAM is the process applied in robotics for building and updating maps and position of an unknown environment [18, 19]. There are different SLAM algorithms which rely in the probabilistic calculation for robot mapping and positioning [19]. The first approach is the utilization of Kalman and extended Kalman Filters [20], which recursively update the previous position estimations, based on a posterior sensor-based measurements. Another approach is graph-based SLAM algorithms which solves some weakness presented by filter-based SLAM. This, by using nodes which represent robot poses and its connection to other nodes (measured by sensors). An edge between two poses represent a spatial constrain between two connected poses. Based on the constructed graph, it is necessary to find the node configuration that is more consistent with the sensor measurements, and that best satisfies the constrains [21]. Most of the 2D SLAM systems found in the literature, use laser scanners as sensors [22]. One of the most widely known 2D SLAM implementations is Google's Cartographer, which details about the local/global approach can be found in [23]. In addition, Kohlbrecher, et al. [24] proposed extremely accurate and reliable 2D SLAM algorithms. Nevertheless, the main drawback of these systems is high amounts of memory usage, which can make them impractical for deployment on lightweight UAVs using cheap onboard computing.

## C. Flight controller and tuning

UAVs systems have both remote controlled and autopilot flight controlling modes. Nevertheless, in the majority of cases UAV systems are flown without assistance of human operators [25]. The most sparse open source options for flight controllers are ArduPilot and PX4 [26]. Despite autopilot utilization, several authors conduct extensive tuning or system identification in order to improve the performance of commercial or open-source controllers. In regard, Saengphet, et al. [27] proposed a detailed procedure to tune the PID parameters in the PX4 controller. The procedure consists of collecting data via piloted test flights and tuning the PID values. The authors claim some improvement in robustness over heuristic methods such as Ziegler- Nichols.

## D. Navigation

Autonomous navigation of mobile robots restricted to 2D planes is a well-researched area due to its extensive applications in ground based robots [28]. On the other hand, less work has been done in relation to UAVs operating at a fixed height. The best navigation results have been obtained by combining global approaches (which do not consider local movement restrictions of the robot) with local approaches (which directly simulate the robot's motion in the immediate future) [28]. Common assumptions in global navigation impose a static environment, and a circular robot footprint to improve computational efficiency [29]. Related with this, several algorithms have been proposed to achieve this task such as the simple implementation of Dijkstra algorithm and A* [30, 31] to more complex genetic algorithms and neural networks [32].

## III. SYSTEM DESIGN

In this study, fixed base-station approach to autonomous flight was implemented. The UAV system was assembled including the components detailed in section II. The selected UAV platform was the Erle Copter from Erle Robotics Company considering as major criteria the combination between system cost, API access, payload limitations and ROS integration. UAV collected data about its surroundings through a planar LiDAR scanner and an ultrasonic sensor for measuring height above the floor (Figure 1). This data was sent to a fixed base station, which performs SLAM and navigation tasks.

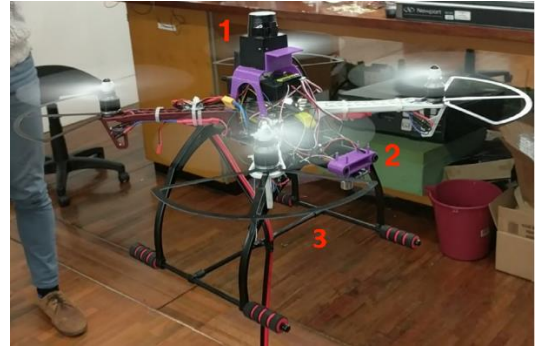

Fig. : Assembled UAV and implemented components.
(1) LiDAR scanner (2) Utrasonic sensor (3) Erle copter UAV

The output of this was an estimate of the UAV's current position and attitude, as well as a target set-point for the UAV to fly towards. This data was transmitted to the UAV, where a modified PX4 flight controller fuses the offboard estimate with onboard inertial sensors operating at faster rates, and issues commands to the actuators to move towards the setpoint. This design is shown graphically in Figure 2.

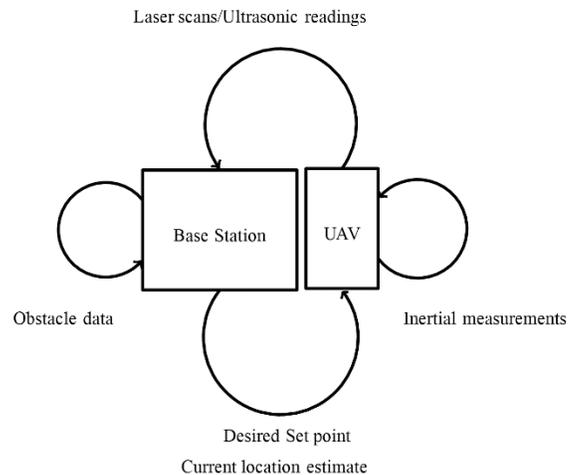

Fig. 2: Diagram of the data transmission protocol implemented in UAV system

To integrate the technologies and algorithms, the Robot Operating System (Kinetic version) was used. The UAV controller was implemented on a Raspberry Pi running the

Raspbian operating system using the PREEMPT_RT patch[3]. Specifications of the software and hardware used are shown in Figure 3. A summary of the 'off-the-shelf' software and hardware used is provided as follows:

*A. Hardware*
- Hokuyo URG-04LX-UG01 laser scanner (5.6 m max range, 240° FOV)
- Raspberry Pi 3B for flight control and onboard computing
- Erle-copter kit for UAV frame/rotors and additional sensors [4]
- SR-04 Ultrasonic sensor for height measurements
- 3D printed mounts for additional sensors
- 650 W, 12 V power supply for tethered flight.

*B. Software*
- Cartographer SLAM algorithm
- PX4 flight controller in standard X configuration
- ROS computing platform
- MAVLink/MAVROS
- Application specific software (drivers, navigation, and signal transforms for integration)

The aim was to combine and evaluate existing technologies which can achieve autonomous indoor flight along with accurately mapping dimensions of different objects in an indoor environment in a 2D plan. The primary contribution was in overall system design and integration. Also, drivers for PX4 to interface with sensors/PWM outputs specific the system configuration were developed. Finally, a customized navigation algorithm was also implemented, in particular by replacing the local planner in the default ROS navigation stack with a class more suitable to control of a UAV by issuing set-points over a wireless link.

## IV. NAVIGATION SYSTEM

Local navigation aims to solve the problem of how to best advance a robot along a given 'global' trajectory, given obstacles in the immediate vicinity, physical limitations and current inertial state of the robot.

*A. ROS implementation*

While the ROS platform has an implementation of local navigation for holonomic ground based robots, it is not appropriate for UAVs since the control spaces do not correspond. Nevertheless, the approach taken by the ROS local planner is instructive. This implementation samples all possible directives which can be given to the robot (its control space) and simulates the trajectories which this would result in.

The trajectories are then allocated costs (denoted as $J$), with the lowest cost trajectory being accepted. It is immediately clear why this simulation approach gives rise to the division between 'local' and 'global' navigation it would not be feasible to simulate all possible trajectories on a global scale, so a less demanding search strategy is used for long-term planning.

The cost is a sum of functions which are formulated to bring about desired behavior such that $J = J_1 + J_2 + ... + J_n$. For example, $J_1$ may represent a desire to avoid obstacles, $J_2$ may represent a desire to progress the robot along the global plan, and $J_3$ may represent a desire for the robot to be facing forward as it traverses the global plan. In the ROS implementation, costs are generally represented by decaying exponentials of the form $J_n(x) = Ae^{-bx}$. This convention was followed, as it generally provided expected behavior in the face of competing priorities.

*B. Adaptation for URSA*

The formulation of the navigation problem outlined above can be adapted to UAVs (or any robot) by undertaking two steps:
1) Appropriately sampling the control space of the UAV and simulation of trajectories
2) Selecting $J_n$ to provide the desired behaviour

A positional set point for the control space was adopted since it is likely to be robust to latency between the navigation/SLAM algorithms and the PX4. A simplified trajectory simulation approach was adopted, which did not model the existing inertia of the UAV. This was justified due to the low velocities at which the UAV was tested.

Sampling the control space was achieved by the following procedure:
1) All points on the global plan within 3 m of URSA were added to the sample set.
2) For each point in (1), two points were chosen at random distances to the left and right. These points were also added to the sample set

The result of this sampling strategy was a 'corridor' of around 100 samples around the global plan. The goal of additional random sampling was to avoid dynamic obstacles, and also allow discovery of better trajectories which provided a greater clearance to existing obstacles.

In relation to $J_n$, following desired behaviors were identified:
1) Turning to face the destination prior to moving.
2) Progressing along the global plan ($J_1$).
3) Maximizing obstacle clearance ($J_2$).
4) Thresholding the setpoint, so it would only change to achieve significant improvements in the above goals ($J_3$)

---

[3]https://rt.wiki.kernel.org/index.php/Main_Page
[4]http://www.erlerobotics.com

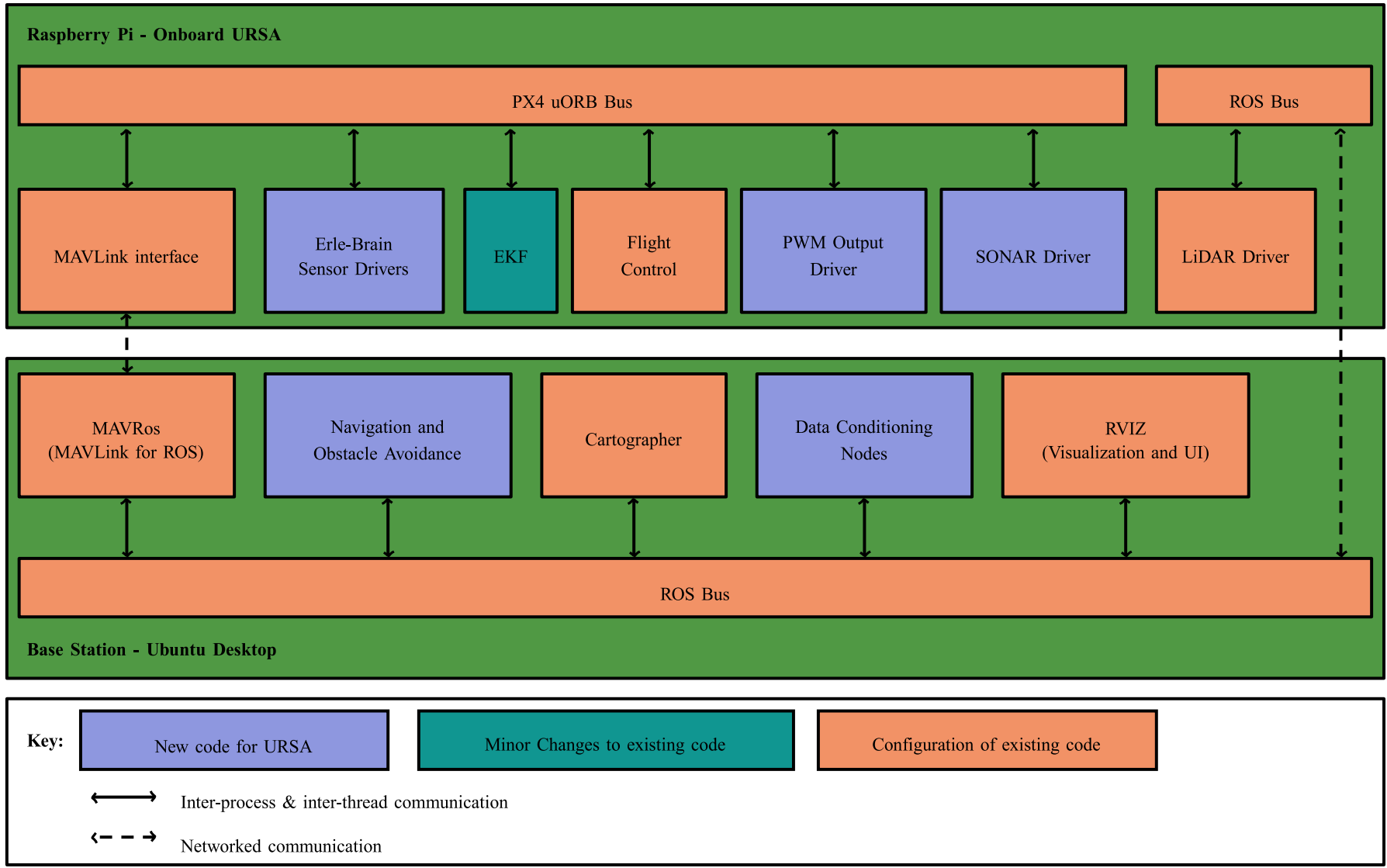

Fig. 3: Software and Hardware architecture implemented in the UAV system

The first of these was achieved by heuristics in the navigation algorithm, and therefore does not have an associated cost function.

*C. Progression/obstacle margin tradeoff*

The tradeoff between path progression ($J_2$) and obstacle avoidance ($J_1$) can be demonstrated using a simplified example as follows:

$$J = J_1 + J_2$$
$$J_1 = Ae^{-c\gamma}$$
$$J_2 = Be^{-d\lambda}$$

Where $A$, $B$, $\gamma$ and $\lambda$ represent free parameters, $c$ is the shortest distance between a trajectory and an obstacle, and $d$ is the distance along the global plan. Intuitively, $J$ favours large $c$ and $d$. However, in many cases, large $c$ implies small $d$ and vice-versa. It then becomes a case of tuning the free parameters to create desired behavior. Assuming that a global path which is a fixed radius around an obstacle was given to the UAV, a 'full circle' global plan is unlikely to occur in practice; however quarter and semi-circle global plans are regularly observed and represent navigating around a corner. This is shown at Figure 4.

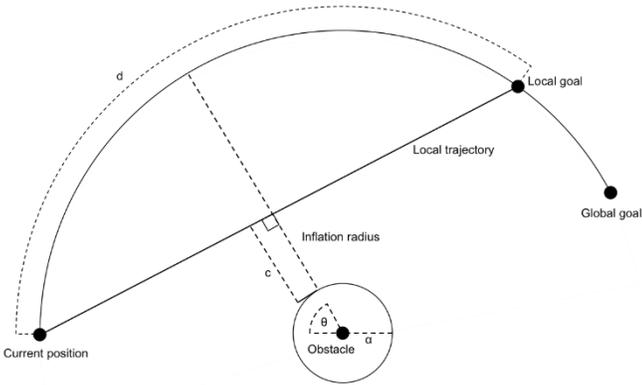

Fig. 4: Model of progression/obstacle margin tradeoff

As can be seen in Figure 4, increasing $d$ implies a monotonic decrease in $c$. The tuning of parameters can therefore be simplified to a question of the desired value of $c$ in this simplified model, where $c$ represents how close the UAV will pass to the obstacle. $A=252$ was chosen for compatibility with ROS costmaps, and $\gamma = 2.5$ so that $J_1$ decays to an insignificant value after around 2 m. This lead to 2 free parameters, which are $B$ and $\lambda$. Figure 5, shows $J = Ae^{-c\gamma} + Be^{d\lambda}$ as a function of $\lambda$ and $c$ for different values of $B$.

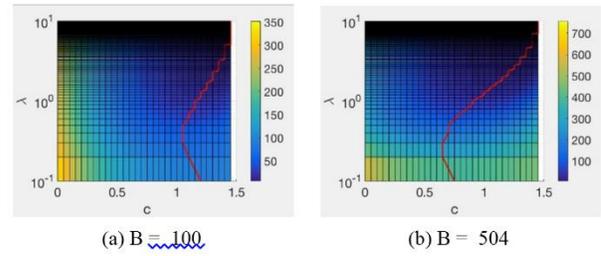

(a) B = 100　　　(b) B = 504

Fig. 5: J as a function of $\lambda$ and $c$ for different values of $B$ (Red line indicates lowest cost $c$ for a given $\lambda$)

On Figure 5, the $c$ results in the lowest cost $J$ for a given $\lambda$ is indicated by the red contour. For very small $\lambda$, the impact of the goal cost function is small as this function decays subtly and almost all goals are penalized equally by $J_2$. Therefore the minima is dictated mostly by $J_1$ and the solution occurs at relatively higher $c$. Conversely, at very high $\lambda$, the goal cost function decays rapidly to zero, making almost no difference to the overall minima, which is again dominated by the obstacle cost. It is only at intermediate values of $\lambda$ that this function appears not to have any effect at all on the minima of the cost function, and solutions which 'cut corners' and reduce $c$ start to be preferred. This indicates that the navigation is more sensitive to global plan progression.

There are two reasons why it is desirable for navigation to be highly sensitive to global plan progression. The first is clear from this model - longer $d$ will result in faster navigation to the global goal and fewer set points along the way. However, the second reason is arguably more important and is not obvious from this simple model. In real examples, URSA is required to approach obstacles in order to progress along the global path (for example, when passing between a doorways). If $J$ is dominated by the obstacle cost, URSA will never select a trajectory that approaches those obstacles. As can be seen from Figure 7, there is a fairly narrow range of $\lambda$ which achieves this ($0.1 < \lambda < 1$). Based on this analysis, $\lambda = 0.4$ and $B = 100$, were chosen which provided 1 m of clearance in the simple case, and high sensitivity to goal progression.

*D. Dynamic obstacle avoidance*

In order to measure $c$ in the above examples, a mechanism is required to find the shortest distance between an obstacle and a trajectory. To this end, the ROS costmap classes was adopted, which allow populating a 2D map with obstacle and propogating cell costs based on distances. To populate the local planner cost map with data was chosen from both Cartographer SLAM, and direct laser scanner data. This means that the local planner will successfully avoid obstacles which are transient and therefore are not mapped by Cartographer.

For the global planner, only Cartographer data was used. This, was combined with regularly updating the global plan at 1 Hz. The result showed that the global plan would reflect the permanent 'structure' of the environment, while the local plan would also reflect transient obstacles. Dynamic obstacles on the global plan could therefore be avoided by the local planner preferring points sampled randomly to the left and right of these obstacles.

## V. EXPERIMENTS

A number of real-life flight experiments were undertaken in a laboratory environment. These experiments fall into two categories: mapping experiments and navigation experiments. For mapping experiments, the goal was to verify that the map generated by a planar laser scanner on a UAV in a crowded indoor environment were reasonably accurate. For navigation experiments, the goal was to test a number of flight patterns and verify that obstacles were correctly avoided.

### A. Mapping

URSA was deployed in a small room which had previously been measured by a tape measure. The test was undertaken in the same room at two different heights: 40 cm and 80 cm. The room was intentionally configured to have certain measurable obstacles (objects). Figure 6, shows photographs of the test environment at 40 cm and 80 cm, respectively. In each case, URSA was instructed to rotate and move around the room until a measurable map was generated. URSA was then instructed to land and the map generated by URSA was compared to the previous tape measurements collected.

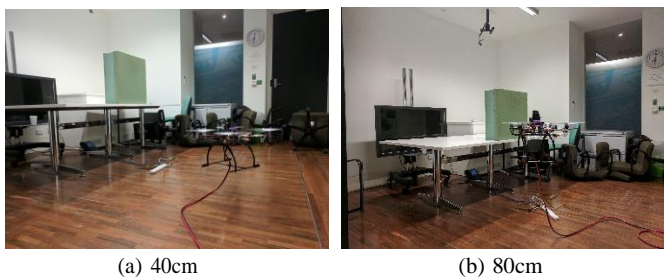

(a) 40cm      (b) 80cm

Fig. 6: Experimental environment setup for different URSA flight altitudes.

Figure 7, shows the 2D map output constructed from sequential laser scans at 40 cm and 80 cm. Table I, summarizes the measurements obtained, showing that there is around a ±2 cm discrepancy between those measurements obtained by hand and those by laser.

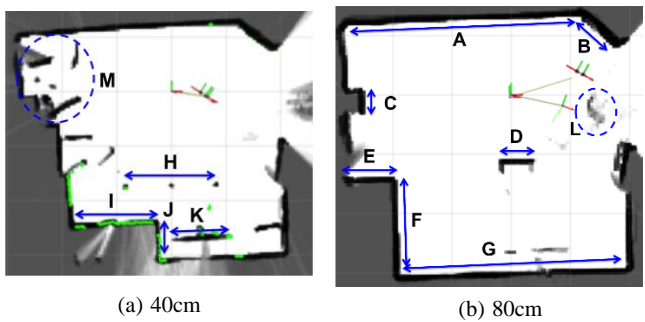

(a) 40cm      (b) 80cm

Fig. 7: Constructed 2D map using SLAM algorithm obtained at different laser scan altitudes.

|   | Tape Measure (m) | UAV-based (m) |
|---|---|---|
| A | 4.65 | 4.67 |
| B | 0.82 | 0.81 |
| C | 0.4  | 0.39 |
| D | 0.56 | 0.54 |
| E | 0.75 | 0.75 |
| F | 1.68 | 1.67 |
| G | 3.9  | 3.94 |
| H | 1.7  | 1.69 |
| I | 1.57 | 1.53 |
| J | 0.73 | 0.75 |
| K | 1.12 | 1.06 |

TABLE I: Comparison between measurements obtained by tape measure and UAV-based measurements.

### B. Navigation

Tests were undertaken to show that URSA was capable of the following fundamental navigation behaviors:

- Avoiding dynamic obstacles
- Turning corners
- Entering a room through a narrow doorway

A video showing each of the experiments conducted is provided with this paper.

1) *Avoiding Dynamic Obstacles:* To test URSA's capability to avoid moving obstacles the following experiment was conducted:

   • The UAV was sent to a goal setpoint from its current position.
   • The path from the current UAV position to the goal setpoint was unobstructed.
   • The path from the current UAV position to the goal setpoint was obstructed by an obstacle (human) while the UAV was navigating.

Success was measured if the UAV was able to generate a trajectory around the obstacle, avoid a collision and reach the original goal setpoint. The location of the dynamic obstacle was chosen such that a path still existed between the UAV and the goal within reasonable margins of obstacles. Figure 8, shows the real-time output to RVIZ when the experiment was conducted.

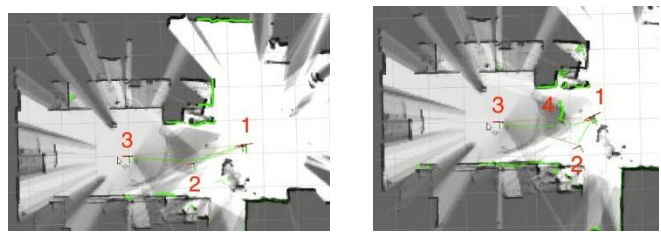

(a) Generated trajectory to goal      (b) Re-generated trajectory around obstacle

Fig. 8: UAV trajectories calculated before and after dynamic obstacles avoiding. (1) UAV location (2) Local trajectory goal (3) Global goal (4) Dynamic obstacle

Figure 8a, shows the first stage of the experiment, where the UAV was asked to generate a trajectory from it's current position to the global goal. During it's progression along the trajectory an obstacle obstructed its path. Figure 8b, shows that a new local trajectory goal was able to be generated to give suitable clearance from the obstacle. The UAV was able to reach the original goal setpoint.

This results shows that the UAV was able to avoid dynamic obstacles in a laboratory setting. A large gap between the UAV and obstacle (1 meter) gave time for the UAV to respond. In subsequent tests this gap should be reduced to find the limitations of the dynamic obstacle avoidance capability.

2) *Turning Corners:* Demonstrating the ability to turn corners illustrates URSA's capability to navigate into previously unseen environments and avoid static obstacles. To evaluate the ability of the UAV to turn corners the following experiment was conducted:

- Place goal setpoint at an unseen location around a corner
- Record any collisions

Success was considered as the ability of the URSA to navigate to the goal setpoint without any collisions.

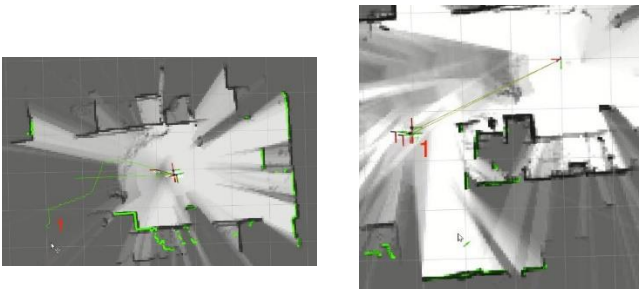

(a) Setpoint placed around corner     (b) UAV reached global point

Fig. 9: Fly trajectory of autonomous UAV avoiding unseen obstacles in turning corners (1) Global setpoint

Figure 9a shows a global goal being placed for the URSA in an unseen area around the corner. While the map had not yet been generated, a goal was able to be placed, indicating an instruction to navigate to an unknown area. In figure 9 the UAV can be seen to have reached the setpoint.

*C. Entering a Room*

A final experiment was conducted to test URSA's ability to fly through narrow passages is shown in Figure 10. The format followed the same structure as described in the previous tests. A global goal was placed inside of a room and URSA success was considered as the ability to reach that global goal without collisions.

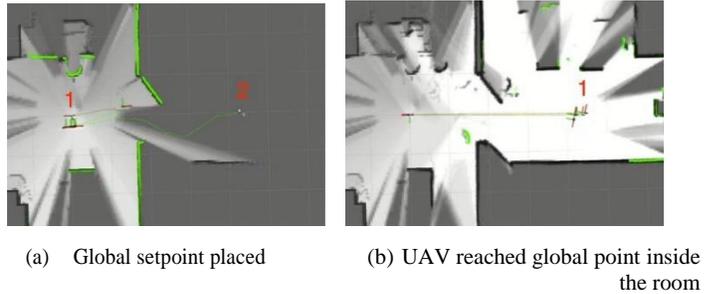

(a) Global setpoint placed     (b) UAV reached global point inside the room

Fig. 10: Flight trajectory of autonomous UAV entering in a room (1) UAV position (2) Global setpoint

Initially, a typical doorway width (0.8m) was used. However, this doorway appeared to be too narrow for the UAV to pass through without any collisions. The gap was then increased until the UAV was able to pass through safely. A gap of 1 m was found to be sufficiently large for the UAV to pass through without collisions.

## VI. CONCLUSION

This study showed the overall design of a prototype system for indoor UAV autonomous flight and navigation with the ability to map an indoor surroundings in real time with an accuracy of 2 cm. The capabilities for a number of mapping and navigation tasks were evaluated. Performance was found to be adequate at low speeds, however there were a number of issues identified at higher speeds. In particular, URSA would often overshoot set points during navigation, leading to unstable behaviour or collisions. A number of conclusions were drawn based on the results obtained for further improvements to reduce or eliminate the above issues:

- Proper simulation of trajectories accounting for inertial effects.
- Tighter integration of the controller, navigation and SLAM algorithms onboard the UAV.

It is also clear that, while the 2D mapping paradigm generally performs quite well for well-structured environments, this approach will not be adequate for more complex or challenging environments. Future solutions are advised to investigate 3D SLAM and navigation approaches.